\documentclass[10pt,twocolumn,letterpaper]{article}

\usepackage{cvpr}
\usepackage{times}
\usepackage{epsfig}
\usepackage{graphicx}
\usepackage{amsmath}
\usepackage{amssymb}
\usepackage{subfigure} 

\usepackage{algorithm}      % algorithm
\usepackage{algpseudocode} % algorithm
\usepackage{caption}        % captionof
\usepackage{array}          % thick column hline
\usepackage{booktabs}       % table style
\usepackage[pagebackref=true,breaklinks=true,letterpaper=true,colorlinks,bookmarks=false]{hyperref}

\cvprfinalcopy % *** Uncomment this line for the final submission

\def\cvprPaperID{5381} % *** Enter the CVPR Paper ID here

\ifcvprfinal\fi
\begin{document}

%%%%%%%%% TITLE
\title{Adaptive Routing Between Capsules}

\author{Qiang Ren\\
\and Shaohua Shang\\
\and Lianghua He\\
Department of Computer Science and Technology, Tongji University\\
{\tt\small {\{rqfzpy,shaohuashang,helianghua\}}@tongji.edu.cn}
%\and
}

\maketitle
%\thispagestyle{empty}

%%%%%%%%% ABSTRACT
\begin{abstract}

Capsule network is the most recent exciting advancement in the deep learning field and represents positional information by stacking features into vectors. The dynamic routing algorithm is used in the capsule network, however, there are some disadvantages such as the inability to stack multiple layers and a large amount of computation. In this paper, we propose an adaptive routing algorithm that can solve the problems mentioned above. First, the low-layer capsules adaptively adjust their direction and length in the routing algorithm and removing the influence of the coupling coefficient on the gradient propagation, so that the network can work when stacked in multiple layers. Then, the iterative process of routing is simplified to reduce the amount of computation and we introduce the gradient coefficient $\lambda$. Further, we tested the performance of our proposed adaptive routing algorithm on CIFAR10, Fashion-MNIST, SVHN and MNIST, while achieving better results than the dynamic routing algorithm.
\end{abstract}

%%%%%%%%% BODY TEXT
\section{Introduction}
%在近几年，深度学习特别是卷积神经网络的的研究取得了极大的成功，在多种计算机视觉的任务上取得了突破性的结果。在卷积神经网络中每个神经元是一个标量，所以它无法在训练中学习到神经元之间的关系，在人的大脑中，神经元不仅仅是单独工作的，而是相互协作，共同完成各种人类的认知活动。针对卷积神经网络的这一缺点，hitton提出了‘胶囊’这一概念\cite{DBLP:conf/icann/HintonKW11}。胶囊是一组神经元的组合，将特征图中的特征（神经元）堆叠成向量（胶囊），使得神经网络在训练的时候不仅考虑到特征，也考虑到特征之间的关系。动态路由算法的出现使‘胶囊’这一思想得以实现\cite{DBLP:conf/nips/SabourFH17}，将神经元堆叠成胶囊后，通过动态路由算法学习低层胶囊和高层胶囊之间的耦合系数，从而获得特征部分与整体之间的关系。
%动态路由算法将胶囊网络从理想变为了现实，但是原始的动态路由算法体现出了种种缺点。经过数次迭代训练后，胶囊网络的关键参数耦合系数cij体现出了较大的稀疏性，说明了只有少部分的低层胶囊对分类的判断是有用的，大部分耦合系数的计算都可以省去，而且每一个胶囊都对应多个神经元，在反向传播的过程中，为网络增加了很大的开销。由于耦合系数的稀疏性，大部分梯度流在胶囊层之间的传播会因为耦合系数的存在变得很小，简单的叠加胶囊层会使梯度更小，使得网络无法正常工作。深度神经网络提高性能的一个常用方法是增加网络的深度，比如VGG\cite{DBLP:journals/corr/SimonyanZ14a}、GoogLeNet\cite{DBLP:conf/cvpr/SzegedyLJSRAEVR15}、和ResNet\cite{DBLP:conf/cvpr/HeZRS16}都是改进算法提高网络的深度，不断刷新了 Image Recognition的准确率。如果想提高胶囊网络的性能，必须改进路由算法，将胶囊层深度增加法。

In last few years, deep learning has made breakthroughs in many computer vision tasks, especially convolutional neural networks leading to state-of-the-art performance. In the convolutional neural network, neurons are scalar and unable to learn the complex relationship between neurons. But in the human brain, neurons usually work together rather than work alone. To overcome this shortcoming of convolutional neural networks, Hitton proposed the concept of ``capsule''\cite{DBLP:conf/icann/HintonKW11} that a combination of neurons that stack features (neurons) of the feature map into vectors (capsules). In the capsule network, the model not only considers the attributes of the feature when training, but also takes account of the relationship between the features. The proposed dynamic routing algorithm enables the idea of ​​``capsule'' to be implemented \cite{DBLP:conf/nips/SabourFH17}. After the neurons are stacked into vectors(capsules), the coupling coefficient between the low-layer capsule and the high-layer capsule is learned through a dynamic routing algorithm. The relationship between the partial features and the whole will be obtained.

%A common way to improve performance in deep neural networks is to increase the depth of the network, such as VGG\cite{DBLP:journals/corr/SimonyanZ14a}, GoogLeNet\cite{DBLP:conf/cvpr/SzegedyLJSRAEVR15}, and ResNet\cite{DBLP:conf/cvpr/HeZRS16} are improved algorithms to improve the depth of the network, constantly refreshing the accuracy of Image Recognition. If the number of layers in the capsule network is deepened, the performance of the model will also improve.

Improving the performance of neural networks is a major direction of deep learning research. A common method to improve the performance of deep neural networks is to increase the depth of the network. For example, VGG\cite{DBLP:journals/corr/SimonyanZ14a}, GoogLeNet\cite{DBLP:conf/cvpr/SzegedyLJSRAEVR15}, and ResNet\cite{DBLP:conf/cvpr/HeZRS16} improves the network depth by proposed effective solutions and continuously improves the accuracy of classification of ImageNet\cite{DBLP:conf/cvpr/DengDSLL009}. In capsule networks, in order to improve the performance of the capsule network can be achieved by increasing the number of capsule layers. Rajasegaran \etal\cite{DBLP:conf/cvpr/RajasegaranJJJS19} have tried and achieved impressive results in this research direction. However, the dynamic routing algorithm proposed by Sabour \etal\cite{DBLP:conf/nips/SabourFH17} cannot simply increase the number of capsule layers in the capsule network.

Dynamic routing algorithm is the method used to learn the relationship between partial features and the whole in a capsule network, but it shows some shortcomings. After several iterations of training, the coupling coefficient $c_{ij}$ of the capsule network shows a large sparsity, indicating that only a small number of low-layer capsules are useful for high-layer capsules. Most coupling coefficient computations are futile, which increases the amount of invalid computation during gradient back-propagation. The sparsity of the coupling coefficients in the dynamic routing algorithm makes most of the gradient flow propagating between the capsule layers very small. If the capsule layer is simply stacked, the gradient in the front layer of the model will become small, so that the model not working. If the interference of the coupling coefficient can be removed during the routing process, the stacked layers can continue to work.

To this end, in this paper, we proposed adaptive routing that a new routing algorithm for capsule networks. Unlike the dynamic routing algorithm, which updates the coupling coefficient at the end of each iteration, our proposed algorithm only updates the low-layer capsule itself at the end of each iteration, which makes the low-layer capsules more "similar" to the high-layer capsules. Since there is no coupling coefficient $c_{ij}$, the propagation of the gradient flow in the capsule network is not suppressed during the routing process, so the gradient can be better transmitted to the layer in front of the model. More specifically, we made the following contributions in this article:

\begin{enumerate}
%\item The motivation proposed by the adaptive routing algorithm, and explains why the dynamic routing algorithm causes the gradient to vanish when iterating multiple layers, causing the capsule network to not work.
\item The motivation proposed by the adaptive routing algorithm and explains why the dynamic routing algorithm causes the gradient vanishing and the capsule network to not work when stacking multiple layers.
\item The adaptive routing algorithm is proposed to overcome the shortcoming that the dynamic routing algorithm will cause the gradient vanishing when stacking multiple layers. The adaptive routing algorithm can stack multiple layers and improve the performance of the capsule network.
\item The iterative process of adaptive routing algorithms can be simplified, and the adaptive routing algorithm without routing process is used. The introduced hyper-parameter $\lambda$ is used instead of the iteration number, which reduces the amount of computation and amplifying the gradient.
\end{enumerate}

%本文的其余部分安排如下：在第2节中，我们讨论与胶囊网络有关的工作，第3节描述了胶囊网络中的梯度流反向传播进行了推导和Adaptive Routing路由算法，第4节显示了我们的实验结果。 最后，第5节总结了论文。
The rest of the paper is organized as follows: In Section 2, we discuss the related work on Capsule Networks, Section 3 describes the motivation and adaptive routing algorithm, Section 4 shows our experimental results. Finally, Section 5 concludes the paper.

\section{Related Work}

The capsule network is a new neural network architecture that stacks traditional scalar neurons into vector neurons called “capsule” neurons\cite{DBLP:conf/icann/HintonKW11} which can store spatial location information of the feature so that it is more in line with the human brain mechanism. The dynamic routing algorithm was proposed by Sabour \etal\cite{DBLP:conf/nips/SabourFH17} that a method learned the coupling relationship between low-layer capsules and high-layer capsules in neural networks so that the capsule network has become a practical model. Then, Hitton \etal\cite{DBLP:conf/iclr/HintonSF18} proposed the EM routing algorithm, which used matrix capsules instead of vector capsules. The EM routing algorithm is used to iteratively learn the coupling coefficient between the low-layer matrix capsule and the high-layer matrix capsule. In the research field of capsule networks, almost researches related to capsule networks are based on these two algorithms.

In this field, there are many great extensions. Lenssen \etal\cite{DBLP:conf/nips/LenssenFL18} proposed a generic routing algorithm that defines the reliable variability and invariance for the capsule network and proved the equal variance of the output pose vector and the output activation. Rajasegaran \etal\cite{DBLP:conf/cvpr/RajasegaranJJJS19} proposed a deep capsule network architecture for the shortcomings of dynamic routing algorithms that cannot simply stack multiple layers. It uses 3D convolution to learn the spatial information between the capsules and the idea of skip connection in the residual network, and the skip connection in the capsule layer allows for a good gradient flow in back-propagation. At the bottom of the network, when skipping connections to more than one layer, a large number of route iterations are used. The 3D convolution is used to generate votes from the capsule tensor for dynamic routing. This helps route a set of localized capsules to a higher layer capsule. Jeong \etal\cite{DBLP:conf/icml/JeongLK19} proposed a new definition method for entities, which deletes the capsules that do not want to be closed and preserves the spatial relationship between low-layer and high-layer entities, and proposed the concepts of building layers and step layers. To capture the relationship between the part and the entire space, another new layer called a ladder layer is introduced, the outputs of which are regressed low-layer capsule outputs from high-layer capsules.
 
These extensions also make a lot of sense. Zhang \etal\cite{DBLP:conf/nips/ZhangEQ18} proposed to use a capsule carrier instead of a neuron activation sample, using a set of capsule subspaces, inputting a feature vector on this set of subspaces, and then using the length of the resulting capsule for the pair scores that fall into different categories. Such a capsule projection network (CapProNet) is trained by learning the orthogonal projection matrix of each capsule subspace and it is shown that each capsule subspace is updated until it contains an input feature vector corresponding to the relevant class. Since the dimension of the capsule subspace is low and an iterative method of estimating the matrix inverse is used, the network can be trained with only a small computational overhead. Ding \etal\cite{DBLP:conf/ijcai/DingWGLW19} divided all capsules into different groups and then performs a group reconstruction routing algorithm to obtain the corresponding advanced capsules. Capsule Max-Pooling is used between the lower and upper layers to prevent overfitting. Li \etal\cite{DBLP:conf/eccv/LiGDOW18} proposed to use two branches to approximate the routing process: one master branch collects the main information from its direct contact in the lower layer, and one auxiliary branch is based on the schema variables encoded in other lower containers to supplement the main information. These two branches communicate in a fast, supervised, and one-time pass compared to previous iterative and unsupervised routing schemes. As a result, the complexity and runtime of the model are reduced dramatically.

\section{Methodology} 
\subsection{Motivation}

\begin{figure*}[htp]
\begin{center}
   \includegraphics[width=0.95\linewidth]{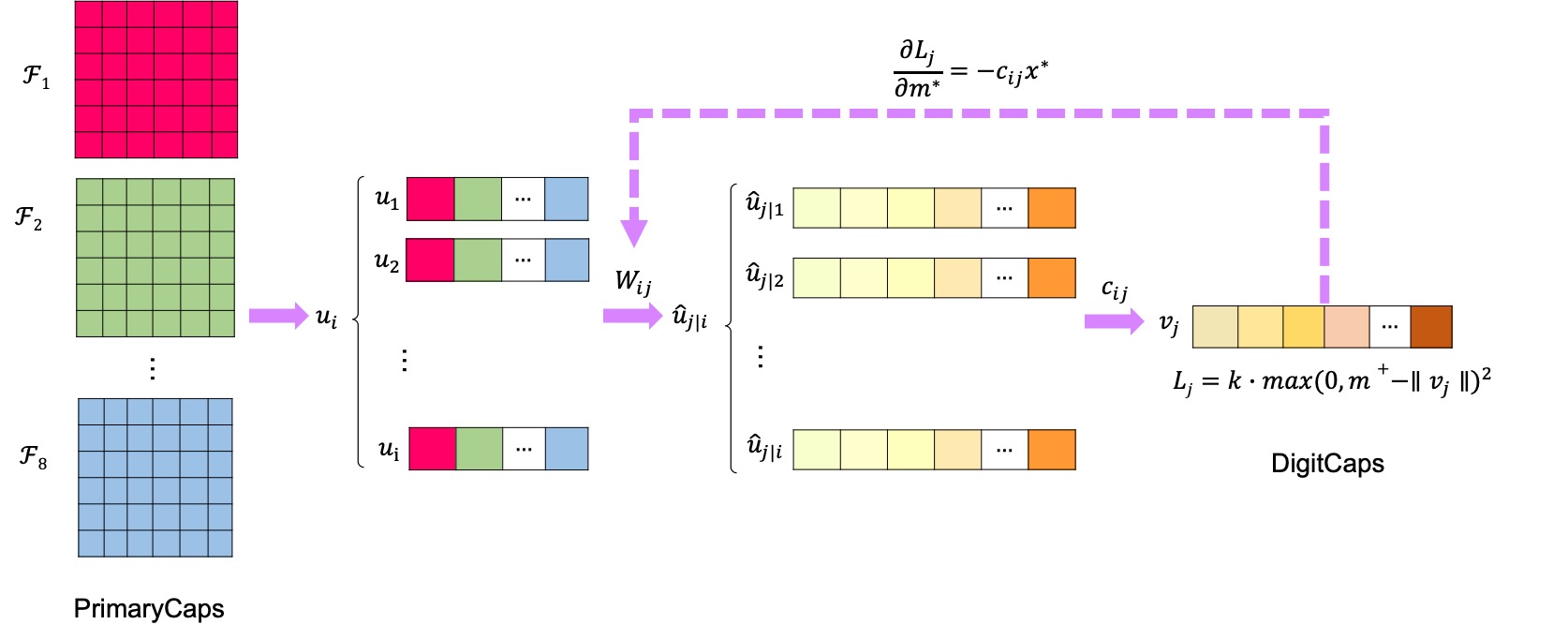}
\end{center}
   \caption{Illustration of forward data flow and backward gradient flow between the PrimaryCaps layer and the DigitCaps layer with the dynamic routing algorithm. The $m^*$ is a parameter in the affine transformation matrix $W_{ij}$, and $x^*$ is the feature associated with $m^*$ in the capsule $u_i$ and on the feature maps $\mathcal F_1$, $\mathcal F_2$, \dots , $\mathcal F_{8}$. The purple solid arrow represents the forward data flow, and the purple dotted arrow represents the backward gradient flow.
}
\label{M1}
\end{figure*}

In the capsule networks used the dynamic routing algorithm, the low-layer capsules learn the ability of affine transformation through the affine transformation matrix $W_{ij}$. Affine transformation matrix is similar to the Transformer Networks proposed by Jaderberg \etal\cite{DBLP:conf/nips/JaderbergSZK15}, enabling the capsule to have the ability to transform, scale, rotate, etc. The capsule network uses the backpropagation algorithm to train the parameters of the affine transformation matrix in the model. The coupling coefficient $c_{ij}$ between the low-layer capsules and the high-layer capsules is iteratively learned by the dynamic routing algorithm. The dynamic routing algorithm outputs the affine-transformed low-layer capsules to the high-layer capsules. During the back-propagation, the coupling coefficient $c_{ij}$ adds weight to the gradient flow. 

%在原始胶囊网络中，特征（标量神经元）堆叠形成胶囊（向量神经元），通过动态路由算法迭代计算低层胶囊和高层胶囊之间的耦合系数cij。梯度流在反向传播的过程中，耦合系数cij相当于对梯度的传播增加一个权重。
%In the capsule network with dynamic routing algorithm, features (scalar neurons) are stacked to form capsules (vector neurons), and the coupling coefficient $c_{ij}$ between the lower layer capsule and the upper layer capsule is iteratively calculated by a dynamic routing algorithm. In the process of backpropagation of the gradient stream, the coupling coefficient $c_{ij}$ is equivalent to adding a weight to the propagation of the gradient.

%For example, after the convolutional neural network, the feature maps in the primarycaps layer are $\textit{X}_{feature}$, $\textit{Y}_{feature}$ and $\textit{Z}_{feature}$. 
Figure \ref{M1} is the illustration of data flow and gradient flow between adjacent capsule layers with the dynamic routing algorithm. Same as the architecture of capsule network proposed by Sabour\cite{DBLP:conf/nips/SabourFH17}, the feature maps in the PrimaryCaps layer are $\mathcal F_1$, $\mathcal F_2$, \dots , $\mathcal F_{256}$. Features on the feature maps as defined in Equation \ref{feaure_map} below: 
\begin{equation}
\begin{aligned}
\mathcal F_n &= (x_1^n,x_2^n,\cdots,x_{36}^n)\\
%\textit{Y} &= (y_1,y_2,\cdots,y_n)\\
%\textit{Z} &= (z_1,z_2,\cdots,z_n)\\
\label{feaure_map}
\end{aligned}
\end{equation}
%低层胶囊uI=（xi,yi,zi……）,经过仿射变换后的低层胶囊为u_hati=（x_hati,y_hati,z_hati……）

Features on the different feature maps are stacked (8 feature maps as a group) and formed into capsules. And all capsules $i$ is in layer $l$ and capsules $j$ is in layer $(l+1)$. Capsules $u_{i}$ in the lower-layer are composed of features on the feature maps $\mathcal F_1$, $\mathcal F_2$, \dots , $\mathcal F_8$ (36 features on each feature map), which are defined according to the Equation \ref{ui} below:
\begin{equation}
\begin{aligned}
u_{i} = (x_i^1, x_i^2, \cdots, x_i^8)
\end{aligned}
\label{ui}
\end{equation}

Affine matrix $W_{ij}$ is defined by Equation \ref{Wij} and transforms the capsule of dimension 8 to the capsule of dimension 16. Therefore, ${ \hat u}_{j|i}$ are obtained by affine transformation of ${u}_{i}$ as defined in Equation \ref{uji} below: 

\begin{equation}
W_{ij} ={
 \left[
 \begin{matrix}
    m_{1}^{1}&  m_{1}^{2}& \cdots    &m_{1}^{16} \\
   m_{2}^{1}&  m_{2}^{2}& \cdots    &m_{2}^{16} \\
   \cdots &  \cdots & \ddots    &\vdots  \\
   m_{8}^{1}&  m_{8}^{2}& \cdots    &m_{8}^{16} \\
  \end{matrix}
  \right]}
\label{Wij}
\end{equation}

\begin{equation}
\begin{aligned}
{\hat u}_{j|i} &= { u}_{i} W_{ij}
\end{aligned}
\label{uji}
\end{equation}

Calculate the weighted sum of ${\hat u}_{j|i}$ and the coupling coefficient ${c_{ij}}$ to get ${ v}_j$ as described in the Equation \ref{vj} below:
\begin{equation}
\begin{aligned}
v_j &= \sum_i {c_{ij}}{\hat u}_{j|i}
\end{aligned}
\label{vj}
\end{equation}

The loss function of the correct category in the capsule network as in Equation \ref{loss} ($m^{+}=0.9$, $k=0.5$) below:

\begin{equation}
\begin{aligned}
L_j &= k\cdot max(0, m^{+} - ||{v}_j||)^2 
%&+ (1-T_j) \cdot max(0, ||{ v}_j|| - m^{-})^2
\end{aligned}
\label{loss}
\end{equation}

%From the Equation \ref{vj}, we can obtain ${ v}_j$'s first neuron ${ v}_j^{(1)}$ as below:
%\begin{equation}
%\begin{aligned}
%{ v}_j^{(1)} &=\sum_i {c_{ij}} { \hat u}_{j|i}^{(1)}\\
%&=\sum_i {c_{ij}}(x_i \times m_{11}^{(i)}+y_i \times m_{21}^{(i)}+z_i \times m_{31}^{(i)})\\
%&=\sum_i ({c_{ij}}m_{11}^{(i)}x_i +{c_{ij}}m_{11}^{(i)}y_i +{c_{ij}}m_{11}^{(i)}z_i )
%\end{aligned}
%\label{vj1}
%\end{equation}

It can be obtained from Equation \ref{loss} that the loss of the capsule networks are related to the length of the capsule ${ v}_j$ and values of the capsule. And $m$ is the parameter in the affine transformation matrix $W_{ij}$, which is learned by the back-propagation algorithm. And $c_{ij}$ is the coupling coefficient, which is learned by iterative calculation of dynamic routing.

%Loss of the first neuron of ${ v}_j$ as described in Equation \ref{vj1_loss}:

%\begin{equation}
%\begin{aligned}
%L_j^{(1)} &= \frac{\partial L_j}{\partial { v}_j^{(1)}}\\
%&=-{ v}_j^{(1)}
%\end{aligned}
%\label{vj1_loss}
%\end{equation}

When the gradient flows through the adjacent capsule layers, the result is as below:
\begin{equation}
\begin{aligned}
\frac{\partial L_j}{\partial m^*}&=\frac{\partial L_j}{\partial { v}_j}\cdot\frac{\partial { v}_j}{\partial m^*}\\
%&=-\frac{\partial { v}_j}{\partial m^*}\\
&=-\frac{\partial \sum_i {c_{ij}}u_iW_{ij}}{\partial m^*}\\
%&=-\frac{\partial \sum_i ({c_{ij}}mu_i)}{\partial m}\\
%&=-{c_{ij}}u_i\\
&=-{c_{ij}}x^*
\end{aligned}
\label{gradient_wij_m11}
\end{equation}

%Similarly, we can obtain the below:
%\begin{equation}
%\begin{aligned}
%\frac{\partial (L_j^{(1)})}{\partial m_{21}^{(i)}}&=-{c_{ij}}y_i
%\end{aligned}
%\label{gradient_wij_m21}
%\end{equation}
%
%\begin{equation}
%\begin{aligned}
%\frac{\partial (L_j^{(1)})}{\partial m_{31}^{(i)}}&=-{c_{ij}}z_i
%\end{aligned}
%\label{gradient_wij_m31}
%\end{equation}
In the Equation \ref{gradient_wij_m11}, the $m^*$ is a parameter in the affine transformation matrix $W_{ij}$, and $x^*$ is the feature associated with $m^*$ in the capsule $u_i$ and on the feature maps $\mathcal F_1$, $\mathcal F_2$, \dots , $\mathcal F_{8}$. The values of the gradient in back-propagation will be affected by the coupling coefficient $c_{ij}$.

\begin{figure}[htp]
\begin{center}
   \includegraphics[width=0.95\linewidth]{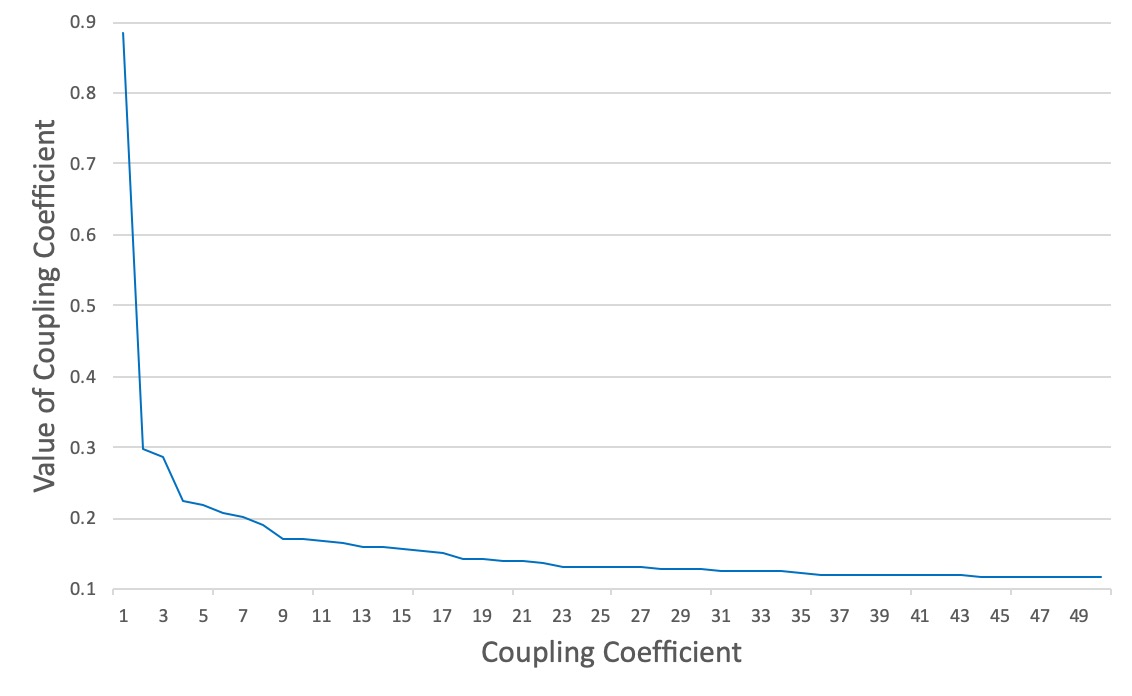}
\end{center}
   \caption{50 maximum values of coupling coefficient $c_{ij}$ in the low-layer predicting the correct digit capsule.}
\label{P1}
\end{figure}

 From the Figure \ref{P1}, the coupling coefficient ${c_{ij}}$  obtained by the dynamic routing algorithm is mostly close to 0.1 and even smaller\cite{DBLP:conf/icml/JeongLK19}. When the capsule networks is stacked in multiple capsule layers, the presence of ${c_{ij}}$  will make the gradient value smaller which affects the learning of the parameters of the front layer and makes the capsule networks not working.

\begin{figure}[htp]
\begin{center}
   \includegraphics[width=0.95\linewidth]{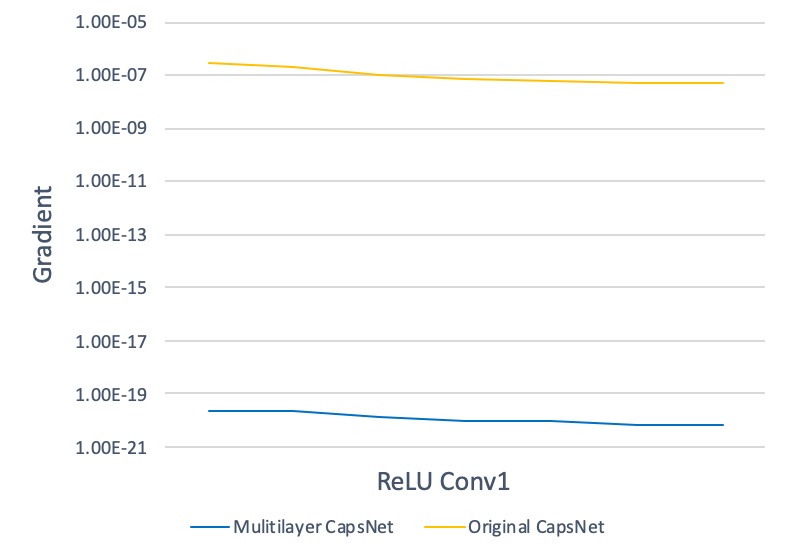}
\end{center}
   \caption{Range of gradients in ReLU Conv1 layer using dynamic routing algorithm.}
   
\label{G1}
\end{figure}
From the Figure \ref{G1}, we compared the range of gradients in the ReLU Conv1 layer in the original capsule network(used dynamic routing algorithm in only two capsule layers) and multiple capsule network. It turns out that in the front layer of the multiple capsule networks, the gradient value is too small for the network to work.

In summary, the loss is related to the length of capsule ${ v}_j$. In the process of gradient back-propagation, the value of ${c_{ij}}$ is close to 0.1 and even smaller, causing the gradient vanishing and making the capsule network not working. If coupling coefficient $c_{ij}$ does not participate in routing iterations, the capsule network will continue to work with multiple capsule layers.

%\subsection{How To Improved Performance With Capsule Networks}
%因为路由过程的存在，所以梯度在胶囊层之间传播相对于传统的神经网络传播会受到一定的影响，深度学习一个提高网络性能的方法就是扩大网络的深度和广度，对于胶囊网络来说，如果有足够大的内存，每一层中胶囊的个数和胶囊的维度比较好扩充，但是对于网络深度的提升是困难的。要解决好梯度的正常传播和控制好计算量，

\subsection{Adaptive Routing}
%动态路由算法是Sabour为了计算相邻胶囊层对应的胶囊之间的耦合系数所提出的。在相邻胶囊层之间，低层的胶囊神经元经过仿射变换后，已经具备了各种空间变换的能力，每个胶囊神经元可以代表一个特征或者一组特征的集合。

%The dynamic routing algorithm is proposed by Sabour to calculate the coupling coefficient of the capsule between adjacent capsule layers\cite{DBLP:conf/nips/SabourFH17}. Between adjacent capsule layers, the capsule neurons in the lower layer after the affine transformation , there is already having the ability to transform, scale, rotate, etc. And each capsule neuron can represent a feature or a collection of features.
In order to overcome the shortcomings of the coupling coefficient $c_{ij}$ in the capsule network. We proposed the adaptive routing algorithm that does not involve parameter training in the route iteration process.

\begin{figure*}[htp]
\begin{center}
   \includegraphics[width=0.95\linewidth]{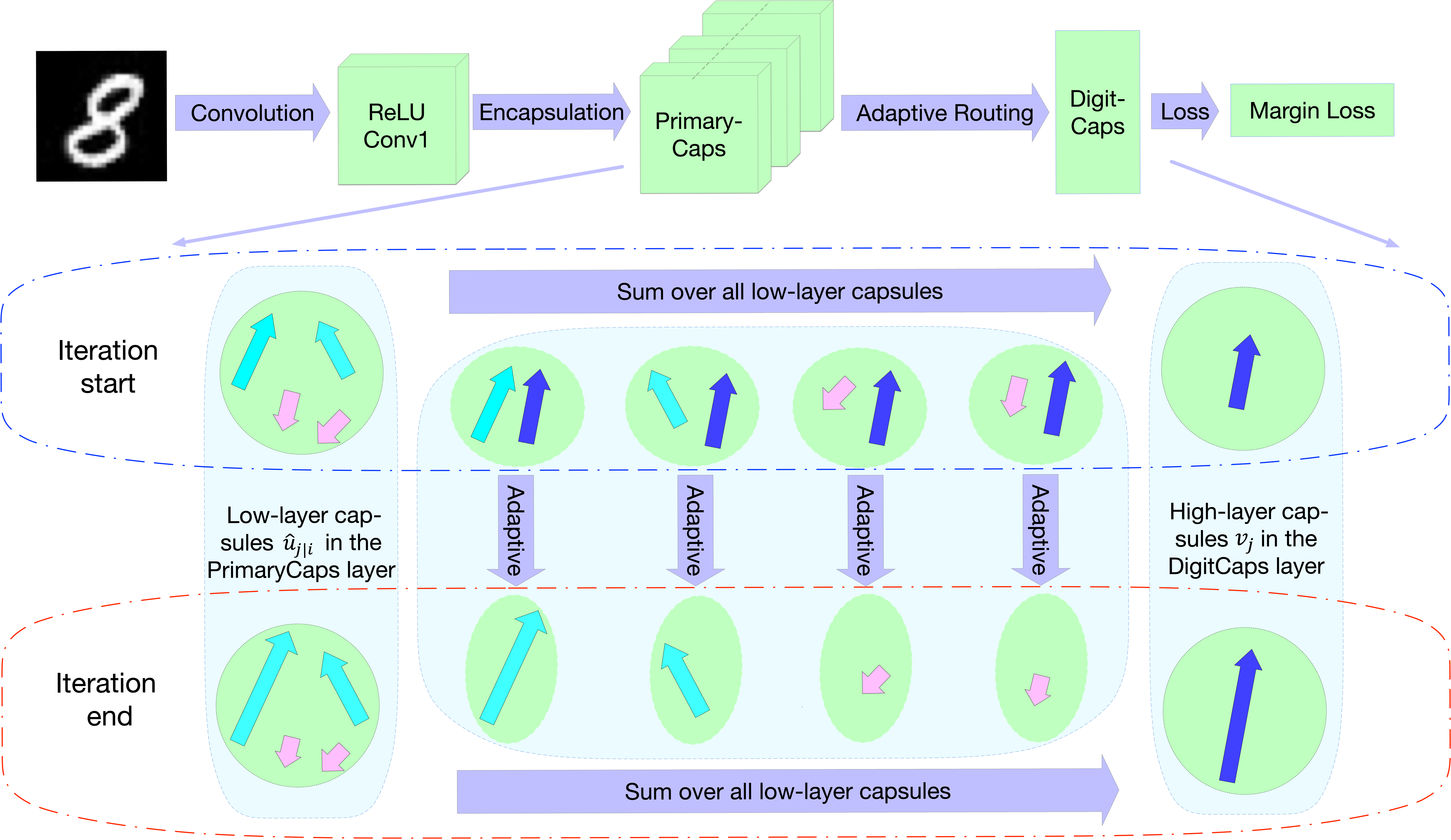}
\end{center}
   \caption{The architecture of the capsule network with the adaptive routing algorithm. This figure is the details of the iterative process of the adaptive routing algorithm. The light blue capsule in the low-layer is similar to the blue capsule in the high-layer, so its length becomes longer after iteration, and the lavender capsule in the low-layer is opposite to the blue capsule in the high-layer, so its length becomes shorter after iteration. During the iterative process, the capsules in the low-layer move adaptively towards the direction of the capsule in the high-layer.}
   \label{M2}
\end{figure*}

In the capsule networks, the direction of the high-layer capsule is close to the maximum direction of the low-layer capsule length, if the coupling coefficient $c_{ij}$ is removed, all the low-layer capsules are directly summed after affine transformation as described in Equation \ref{adap_sj} below:

\begin{equation}
\begin{aligned}
{ s}_j =\sum_i{\hat u}_{j|i}
\end{aligned}
\label{adap_sj}
\end{equation}

Squeeze ${s}_j$ using the activation function (\texttt{squash}), then we can obtain ${v}_j $(same direction as sj) as in Equation \ref{adap_vj}:

\begin{equation}
\begin{aligned}
{v}_j = \texttt{squash}({ s}_j)
\end{aligned}
\label{adap_vj}
\end{equation}

From the Figure \ref{M2}, the direction of the corresponding high-layer capsule ${ v}_j$ is the same as that of the longer capsule in the lower layer. The purpose of the dynamic routing algorithm is that if the low-layer capsule and the corresponding high-layer capsule have higher similarity, the bigger the coupling coefficient between them after iteration. Thus, we can move the low-layer capsule towards the corresponding high-layer capsules. If the low-layer capsule and the corresponding high-layer capsule have higher similarity, then the new ${\hat{u}}_{j|i}$ moved toward the corresponding high-layer capsule, enhanced directionality based on the original ${\hat{u}}_{j|i}$. And if the low-layer capsule and the corresponding high-layer capsule have lower similarity, the new ${\hat{u}}_{j|i}$ also moved toward the corresponding high-layer capsule, reduced directionality based on the original ${\hat{u}}_{j|i}$. The ${\hat{u}}_{j|i}$  adaptive update process is as defined in Equation \ref{adaptive_update_process} below:
\begin{equation}
\begin{aligned}
{ \hat{u}}_{j|i} = {v}_j + {\hat{u}}_{j|i}
\end{aligned}
\label{adaptive_update_process}
\end{equation}

%Different from the traditional dynamic routing algorithm, our update here only considers the direction of their linear sum, while dynamic routing considers parameters such as modulus length and angle. Our algorithm makes the orientation of the high-layer capsules more accurate because our algorithm is more directional than the dynamic routing algorithm after the same number of iterations.
%\begin{figure}[t]
%\begin{center}
%   \includegraphics[width=0.95\linewidth]{2}
%\end{center}
%   \caption{Adaptive routing algorithm update process.}
%\label{T2}
%\end{figure}

 The adaptive routing algorithm can be described as Algorithm \ref{adap1}.
 \begin{algorithm}
\caption{{Adaptive algorithm.}\label{adap1}}
\begin{algorithmic}[1]
\Procedure{Routing}{${\hat{u}}_{j|i}$, $r$, $l$}
\State capsule $i$ in layer $l$ and capsule $j$ in layer $(l+1)$
\For{$r$ iterations}
\State  ${ s}_j \gets \sum_i{\hat u}_{j|i}$
\State  ${ v}_j \gets \texttt{squash}({s}_j)$ 
\State  ${ \hat{u}}_{j|i} \gets { v}_j + { \hat{u}}_{j|i}$
\EndFor
\State \Return ${ v}_j$
\EndProcedure
\end{algorithmic}
\end{algorithm}

\begin{figure}[htp]
\begin{center}
   \includegraphics[width=0.49\linewidth]{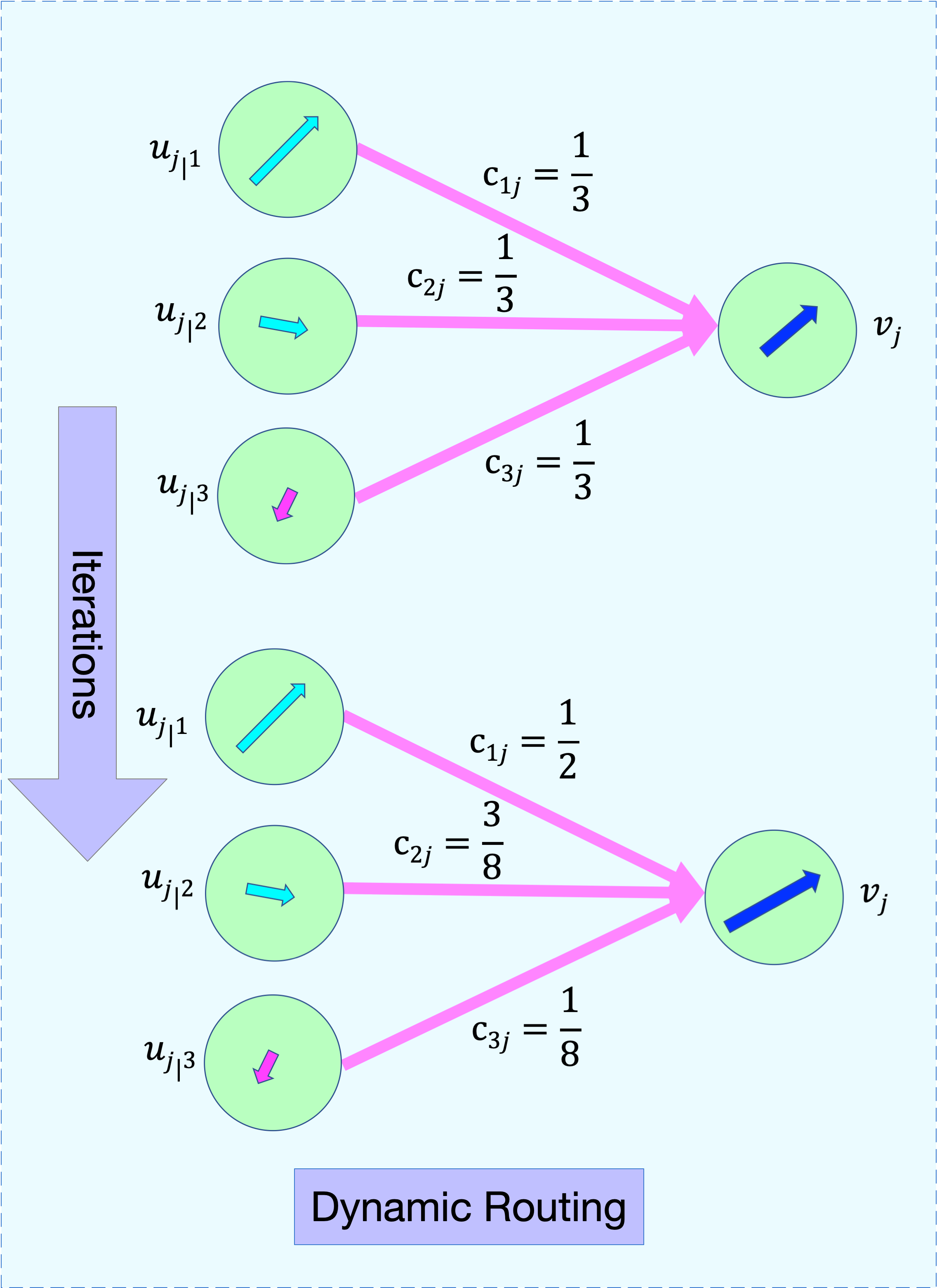}
   \includegraphics[width=0.49\linewidth]{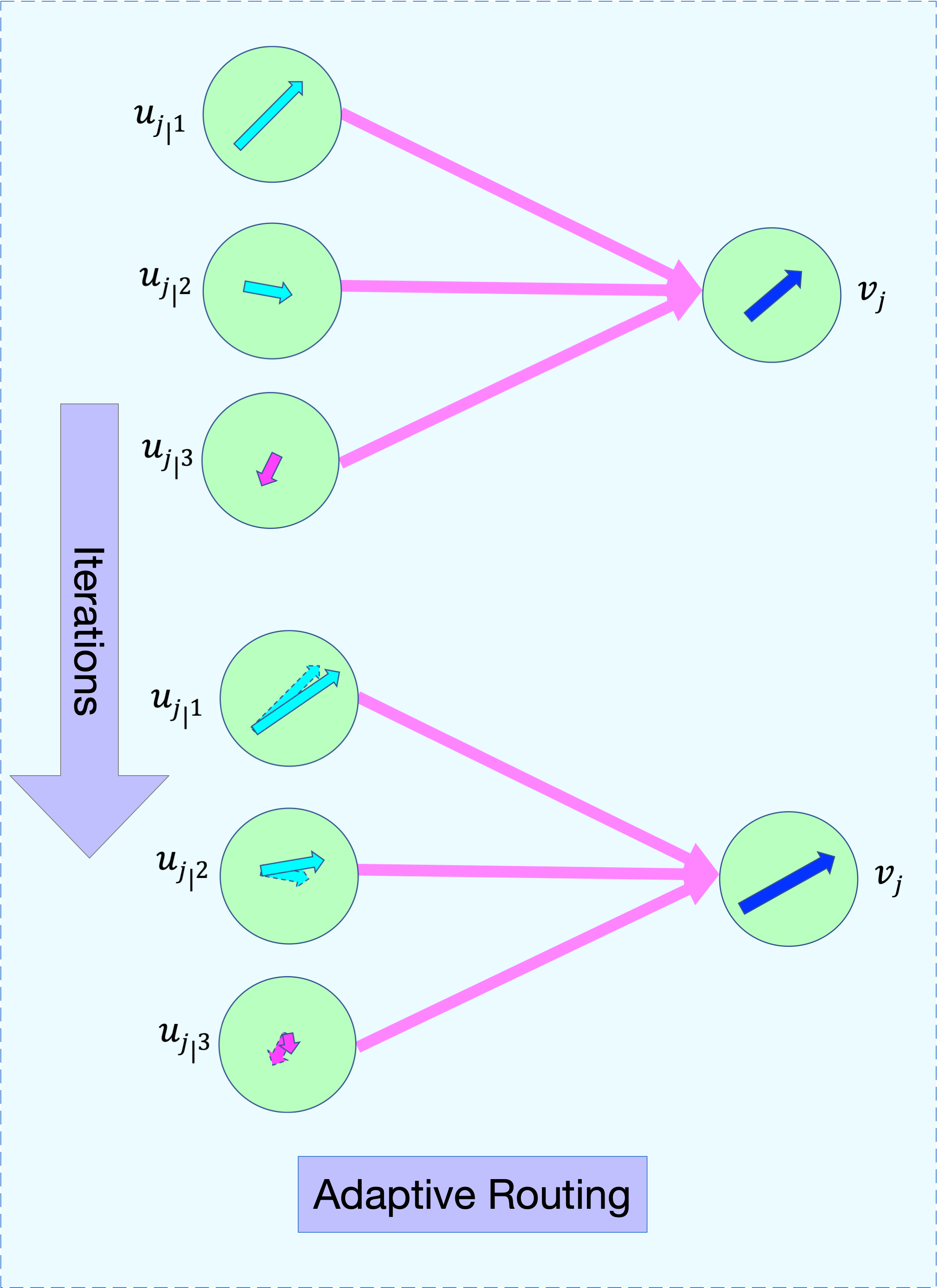}
   \caption{Two figures are comparisons of the iterative process of the dynamic routing algorithm and the adaptive routing algorithm.}
   \label{DA}
\end{center}
\end{figure}

%在原始的胶囊网络结构中uij是经过仿射变换后的低层的胶囊神经元，当动态路由算法迭代开始时，每个低层的胶囊神经元对于高层的相对应的神经元的耦合系数cij都是相等的，所以在第一轮迭代后，经过加权线性求和，如果低层胶囊的模长越大，则它的方向越相似于对应的高层胶囊的方向。之后每次迭代，根据低层胶囊于相对应的高层胶囊的点积（相似度和模长）去更新权重。经过动态路由算法后，高层胶囊的方向都接近于低层胶囊中模长较长的胶囊的方向，耦合系数的存在相当于随着迭代次数的增加，使它们相似度越大的，权重越大，相似越小的，权重越小。

%The dynamic routing algorithm is proposed by Sabour to calculate the coupling coefficient of the capsule between adjacent capsule layers\cite{DBLP:conf/nips/SabourFH17}. 
In the capsule networks used dynamic routing algorithm, ${\hat{u}}_{j|i}$ is a low-layer capsule neurons after affine transformation. From the dynamic routing in Figure \ref{DA}, when the dynamic routing algorithm starts iterating, the coupling coefficient $c_{ij}$ of each low-layer capsule neuron for the corresponding high-layer neurons is equal. ${\hat{u}}_{j|1}$, ${\hat{u}}_{j|2}$, ${\hat{u}}_{j|3}$  are weighted sum to get $v_j$, and the weights are $c_{1j}$, $c_{2j}$, $c_{3j}$. After the first routing, calculate the weighted sum for overall low-layer capsules. If the length of the low-layer capsule is larger, its direction is more similar to the direction of the corresponding high-layer capsule. After each iteration, the coupling coefficient $c_{ij}$ are updated according to the dot product (similarity and length) of the low-layer capsules and the corresponding high-layer capsules. Update the new weights $c_{1j}$, $c_{2j}$, $c_{3j}$ after iterating through the dynamic routing algorithm. If ${\hat{u}}_{j|1}$, ${\hat{u}}_{j|2}$, ${\hat{u}}_{j|3}$ and $v_j$ are more similar, then $c_ij$ becomes larger after updating. Similarly, $v_j$ becomes larger in the same direction before iteration. After the dynamic routing algorithm, the orientation of the high-layer capsules is close to the direction of the longer length capsules in the low-layer capsules. With the number of iterations increased, if the low-layer capsules are more similar to corresponding high-layer capsules, the coupling coefficient $c_{ij}$ (weight) is larger. On the other hand, the $c_{ij}$ is smaller.

 Similarly, from the adaptive routing in Figure \ref{DA}, when the adaptive routing algorithm starts to iterate, the coupling coefficient $c_{ij}$ of each low-layer capsule neuron for the corresponding high-layer neurons is removed. $u_{j|1}$, $u_{j|2}$, $u_{j|3}$ are summed to get $v_j$. After the first routing, sum overall low-layer capsules. If the length of the low-layer capsule is larger, its direction is more similar to the direction of the corresponding high-layer capsule. After each iteration, move the low-layer capsule $u_{j|1}$, $u_{j|2}$, $u_{j|3}$  to the direction of the high-layer capsule $v_j$. After each iteration, the low-layer capsule $u_{j|1}$, $u_{j|2}$, $u_{j|3}$ will become closer the direction of the high-layer capsule $v_j$. After the process of adaptive routing algorithm, the orientation of the high-layer capsules is close to the direction of the capsules with longer lengths in the low-layer capsules. The low-layer capsules will move adaptively to high-layer capsules increasingly, and high-layer capsules will definitely represent the probability of object presence. Without the influence of the coupling coefficient $c_{ij}$, the same effect as the dynamic routing algorithm can be obtained.
% when the dynamic routing algorithm starts to iterate, $u_1$, $u_2$, $u_3$ are weighted linearly summed to get $v_1$, and the weights are $c_1$, $c_2$, $c_3$. In the Figure\ref{T1}(b), Update the new weights $c_1'$, $c_2'$, $c_3'$ after iterating through the dynamic routing algorithm. If $u_1$, $u_2$, $u_3$ and $v_1$ are more similar, then $c_i$ becomes bigger after updating. Similarly, $v_1'$ becomes larger in the same direction as $v1$.

%既然高层胶囊的方向，接近于低层胶囊模长最大的方向，那么如果去掉耦合系数这一参数，将经过仿射变换后的所有低层胶囊直接线性求和，相对应的高层胶囊的方向也和低层胶囊中模长最长的胶囊的方向相同。因为动态路由的最终目的是使与高层胶囊相似的越大的低层胶囊权重越大，与高层胶囊相似的越小的胶囊权重越小，所以我们可以将低层胶囊与其相对应的高层胶囊分别线性求和，如果低层胶囊和相对应的高层胶囊的相似的较大，那么更新之后新的uij偏向于高层胶囊的变换就大，如果低层胶囊和相对应的高层胶囊的相似度较小，那么更新之后新的uij偏向于高层胶囊的变换就小。与传统动态路由的算法不同的是，我们这里的更新只考虑了它们线性和的方向，而动态路由则考虑了模长与夹角等参数。我们的算法使得高层胶囊的方向可以更加的准确，因为经过相同的迭代次数，我们的算法比动态路由算法的方向性更强。

%因为在路由迭代的时候没有参数的学习，仅仅是在每轮迭代后低层胶囊和高层胶囊进行线性求和，所以当迭代轮数r=1的时候，

%\subsection{Replace the number of iterations with $\lambda$}
\subsection{Introduce the gradient coefficient $\lambda$}

The adaptive routing we proposed does not involve the coupling coefficient ${c_{ij}}$ in the routing process. And we can simplify the training process of adaptive routing. No parameters need to be trained during the route iteration, only the capsules in the lower layer are summed. When the iteration r=1, the training process of adaptive routing as in Equation \ref{r1s}, \ref{r1v}, \ref{r1u} below:
\begin{equation}
\begin{aligned}
{s}^{(r=1)}_j &= \sum_i{\hat u}^{(r=1)}_{j|i}
\end{aligned}
\label{r1s}
\end{equation}
\begin{equation}
\begin{aligned}
{ v}^{(r=1)}_j& = \texttt{squash}({ s}^{(r=1)}_j)\\
 &= \texttt{squash}(\sum_i{ \hat u}^{(r=1)}_{j|i})
\end{aligned}
\label{r1v}
\end{equation}
\begin{equation}
\begin{aligned}
{ \hat{u}}^{(r=2)}_{j|i} &= { v}^{(r=1)}_j + { \hat{u}}^{(r=1)}_{j|i} \\
&= \texttt{squash}({ s}^{(r=1)}_j) +  \hat{u}^{(r=1)}_{j|i}\\
&= \texttt{squash}(\sum_i{ \hat u}^{(r=1)}_{j|i}) +  \hat{u}^{(r=1)}_{j|i}
\end{aligned}
\label{r1u}
\end{equation}

%因此一轮迭代后，路由算法的输出
So after the first iteration, the output of the adaptive routing algorithm as in Equation \ref{r1vj} below:
\begin{equation}
\begin{aligned}
{ v}^{(r=1)}_j &= \texttt{squash}(\sum_i{ \hat u}^{(r=1)}_{j|i})
\end{aligned}
\label{r1vj}
\end{equation}

%更新后的uij为
Combine Equation \ref{r1u} and \ref{r1vj}, the input ${\hat{u}}_{j|i}$ of the second iteration is updated to:
\begin{equation}
\begin{aligned}
{ \hat{u}}^{(r=2)}_{j|i} &= \texttt{squash}(\sum_i{ \hat u}^{(r=1)}_{j|i}) +  \hat{u}^{(r=1)}_{j|i}
\end{aligned}
\end{equation}

%当迭代轮数r=2的时候，
 When the iteration r=2, the training process of adaptive routing as in Equation \ref{r2s}, \ref{r2v}, \ref{r2u} below:

\begin{equation}
\begin{aligned}
{ s}^{(r=2)}_j &= \sum_i{ \hat u}^{(r=2)}_{j|i}
\end{aligned}
\label{r2s}
\end{equation}

\begin{equation}
\begin{aligned}
{ v}^{(r=2)}_j &= \texttt{squash}({ s}^{(r=2)}_j) \\
&= \texttt{squash}(\sum_i{ \hat u}^{(r=2)}_{j|i}) \\
&= \texttt{squash}(\sum_i(\texttt{squash}\sum_i({ \hat u}^{(r=1)}_{j|i}) +  \hat{u}^{(r=1)}_{j|i})) \\
&= \texttt{squash}(\sum_i\texttt{squash}(\sum_i{ \hat u}^{(r=1)}_{j|i}) + \sum_i \hat{u}^{(r=1)}_{j|i}) \\
&\approx \texttt{squash}(\lambda\sum_i \hat{u}^{(r=1)}_{j|i})
\end{aligned}
\label{r2v}
\end{equation}

\begin{equation}
\begin{aligned}
{ \hat{u}}^{(r=3)}_{j|i} &= { v}^{(r=2)}_j + { \hat{u}}^{(r=2)}_{j|i}\\
&\approx  \texttt{squash}(\lambda\sum_i \hat{u}^{(r=1)}_{j|i}) + { \hat{u}}^{(r=2)}_{j|i}
\end{aligned}
\label{r2u}
\end{equation}

The introduction of $\lambda$ indicates that the $\sum_i \hat{u}^{(r=1)}_{j|i}$ is amplified, and its value is close to ${ v}^{(r=2)}_j$ after the activation function.

%因此第二轮迭代后，路由算法的输出

So after the second iteration, the output of the adaptive routing algorithm as in Equation \ref{r2vj} below:
\begin{equation}
\begin{aligned}
{ v}^{(r=2)}_j & \approx \texttt{squash}(\lambda\sum_i \hat{u}^{(r=1)}_{j|i})
\end{aligned}
\label{r2vj}
\end{equation}

%综上，如果迭代次数增加，则lambda会更大，路由算法可改善为

In summary, if the number of iterations increased, the $\lambda$ will be larger and finally get ${v}_j$ as in Equation \ref{fvj} below:
\begin{equation}
\begin{aligned}
{ v}_j & \approx \texttt{squash}(\lambda\sum_i \hat{u}_{j|i})
\end{aligned}
\label{fvj}
\end{equation}

The improved adaptive routing without iteration is described as Algorithm \ref{adap2}.
\begin{algorithm}
\caption{{Adaptive Routing Without Iteration.}\label{adap2}}
\begin{algorithmic}[1]
\Procedure{Routing}{${\hat{u}}_{j|i}$, $r$, $l$}
\State capsule $i$ in layer $l$ and capsule $j$ in layer $(l+1)$
\State  ${ s}_j \gets \sum_i{ \hat u}_{j|i}$
\State  ${ v}_j \gets \texttt{squash}(\lambda{ s}_j)$ 
\State \Return ${ v}_j$
\EndProcedure
\end{algorithmic}
\end{algorithm}

%The adaptive routing we proposed does not involve the coupling coefficient ${c_{ij}}$  in the routing process. And the improved adaptive routing replaces the number of iterations with a hyperparameter $\lambda$.

%From the Figure. We can get the gradient value range of the conv1 layer and the primarycaps layer under the adaptive routing capsule network. It turns out that in the front layer of the adaptive routing capsule network, The range of gradient values is within the range that the neural network can continue to train.

In the adaptive routing algorithm ${ v}_j$ as described in the Equation \ref{r_vj} below:
\begin{equation}
\begin{aligned}
{ v}_j &= \lambda \sum_i{ \hat u}_{j|i}
%&=\lambda\sum_i({ \hat u}_{j|i}^{(1)}, { \hat u}_{j|i}^{(2)}, { \hat u}_{j|i}^{(3)}, { \hat u}_{j|i}^{(4)}, { \hat u}_{j|i}^{(5)}, { \hat u}_{j|i}^{(6)})\\
%&=({ v}_j^{(1)},{ v}_j^{(2)},{ v}_j^{(3)},{ v}_j^{(4)},{ v}_j^{(5)},{ v}_j^{(6)})
\label{r_vj}
\end{aligned}
\end{equation}

%From the Equation\ref{r_vj}, we can get:
%\begin{equation}
%\begin{aligned}
%{ v}_j^{(1)} &=\lambda \sum_i { \hat u}_{j|i}^{(1)}\\
%&=\lambda \sum_i (x_i \times m_{11}^{(i)}+y_i \times m_{21}^{(i)}+z_i \times m_{31}^{(i)})\\
%&=\lambda \sum_i (m_{11}^{(i)}x_i +m_{21}^{(i)}y_i +m_{31}^{(i)}z_i )
%\end{aligned}
%\label{r_vj1}
%\end{equation}

 Combine Equation\ref{r_vj} and \ref{loss}, we will get the gradient flows through the adjacent capsule layers used adaptive routing as below(the meaning of $m^*$ and $x^*$ is equivalent to Equation \ref{gradient_wij_m11}):

\begin{equation}
\begin{aligned}
\frac{\partial L_j}{\partial m^*}&=\frac{\partial L_j}{\partial { v}_j}\cdot\frac{\partial { v}_j}{\partial m^*}\\
&=-\frac{\partial \lambda \sum_i u_iW_{ij}}{\partial m^*}\\
&=-\lambda x^*
\end{aligned}
\label{r_gradient_wij_m11}
\end{equation}

%\begin{equation}
%\begin{aligned}
%\frac{\partial (L_j^{(1)})}{\partial m_{21}^{(i)}}&=-\lambda y_i
%\end{aligned}
%\label{r_gradient_wij_m21}
%\end{equation}
%
%\begin{equation}
%\begin{aligned}
%\frac{\partial (L_j^{(1)})}{\partial m_{31}^{(i)}}&=-\lambda z_i
%\end{aligned}
%\label{r_gradient_wij_m31}
%\end{equation}

%在胶囊层之间的反向传播中的等式中，动态路由算法的梯度系数cij接近于0，对梯度有缩小作用。自适应路由算法中的梯度系数l为一个超参数，通常大于0的正整数，对梯度有放大作用，使得前面层的神经网络在多层胶囊网络时依然可以正常工作
By comparing the Equation \ref{r_gradient_wij_m11} the Equation \ref{gradient_wij_m11} we obtained the improvement of the gradients in the back-propagation between the capsule layers. The gradient coefficient $c_{ij}$ of the dynamic routing algorithm is mostly close to 0.1 or even smaller, which causes the gradient vanishing. The gradient coefficient $\lambda$ of the adaptive routing algorithm is a hyper-parameter, usually a positive integer greater than 1, which amplifies the gradient. 
%So the neural network in the front layer can still work in the multiple capsule layers network.
\begin{figure}[htp]
\begin{center}
   \includegraphics[width=0.95\linewidth]{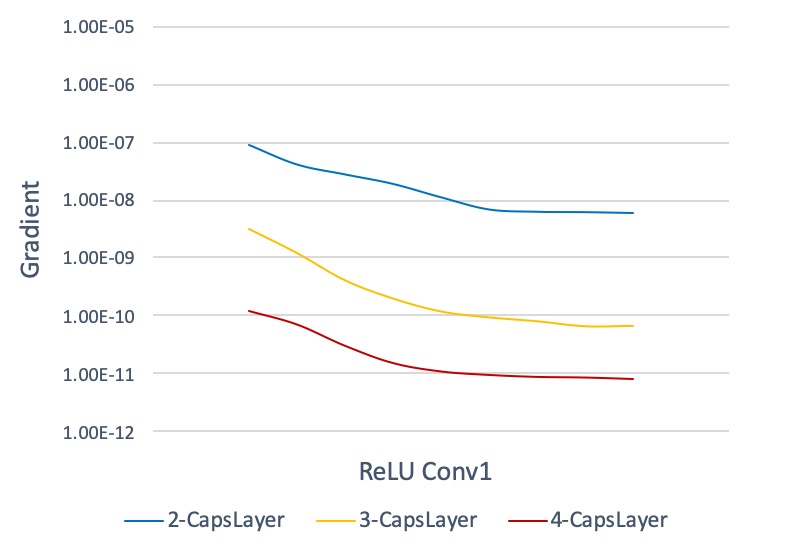}
\end{center}
   \caption{Range of gradients in ReLU Conv1 layer using adaptive routing algorithm.}
\label{G2}
\end{figure}

From the Figure \ref{G2}, we compared the range of gradients in the ReLU Conv1 layer in the multiple capsule layers network(with adaptive routing). Compared with the results of dynamic routing algorithm in the Figure \ref{G1}, it turns out that in the front layer of the multiple capsule layers network, the value of the gradients is larger and the capsule network still continue to work.

The hyper-parameter  $\lambda$ not only inhibits the gradient vanishing to some extent, but also the appropriate $\lambda$ can magnify the gradient and spread the gradient more smoothly to the front of the model.
\section{Experiments} 
%参考的论文baseline，
%姿态学习Group Equivariant Capsule Networks
%基本实验DeepCaps: Going Deeper with Capsule Networks
%Group Reconstruction and Max-Pooling Residual Capsule Network，Ladder Capsule Network，The Multi-Lane Capsule Network，Sparse Unsupervised Capsules Generalize Better，PATH CAPSULE NETWORKS，Multi-layer Dense Capsule Networks
\subsection{Implementation}

We tested our proposed adaptive routing algorithm for classification experiment on several common datasets, MNIST\cite{mnist}, Fashion-MNIST\cite{fashionmnist}, SVHN\cite{svhn} and CIFAR-10\cite{cifar}. For CIFAR-10 and SVHN, we resized the images to $32 \times 32 \times 3$ and shifted by up to $2$ pixels in each direction with zero padding, and there is no other data augmentation/deformation. For other datasets, original image sizes are used throughout our experiments. In the experiment of two capsule layers, we set the number of capsules per layer is [1152, 10] and the same as the dynamic routing algorithm\cite{DBLP:conf/nips/SabourFH17}. And for the experiment of three capsule layers and four capsule layers, the number of capsules per layer we set is [1152, 256, 10] and [1152, 256, 32, 10] respectively.

We used pytorch libraries for the development of experiment. For the training procedure, we used Adam optimizer with an initial learning rate of 0.001, which is reduced $5\%$ after each epochs\cite{DBLP:journals/corr/KingmaB14}. We set the batchsize is 128 that train with 128 images each time. The models were trained on GTX-1080Ti and training 150 epoch for every experiment. All experiments were run three times and the results were averaged. 

\subsection{Classification Results}
We tested our proposed adaptive routing algorithm and dynamic routing algorithm on several benchmark datasets, CIFAR10 \cite{cifar}, SVHN \cite{svhn}, Fashion-MNIST \cite{fashionmnist} and MNIST \cite{mnist}.

\begin{table}[!htb]
\caption{Classification accuracies of dynamic routing algorithm(DRA) and our proposed adaptive routing algorithm(ARA) with the same configuration as two capsule layers.}
\begin{center}
\label{result1}
\begin{tabular}{|l|c|c|c|c|c|}
\hline
Model  & \footnotesize{CIFAR10} & \footnotesize{SVHN} & \footnotesize{F-MNIST} & \footnotesize{MNIST}  \\
\hline
DRA&  76.05\%     & 93.65\%  &   93.02\%      &  99.65\%\\
\hline
ARA& \bf78.41\%     & \bf94.27\%  &   \bf93.07\%      &  99.65\%\\

\hline
\end{tabular}
\end{center}
\vspace*{-0.5cm}
\end{table}

From the Table \ref{result1}, we have obtained the same network configuration and achieved better performance than the dynamic routing algorithm. The routing algorithm between the capsule layers learns the affine transformation of the object and the combination of low-layer capsules and high-layer capsules. Therefore, stacking multiple capsule layers can improve model performance, which can learn more powerful affine transformation capabilities and more complex combinations corresponding adjacent layer capsules.

%\begin{table}[!htb]
%\caption{Classification accuracies of adaptive routing algorithm in different numbers of capsule layers.}
%\begin{center}
%\label{result2}
%\begin{tabular}{|l|c|c|c|c|c|}
%\hline
%Number of layers  & \footnotesize{CIFAR10} & \footnotesize{SVHN} & \footnotesize{F-MNIST}\\
%\hline
%2-layers&  76.05\%     & 93.65\%  &   93.02\% \\
%\hline
%3-layers& 78.50\%     & 94.27\%  &   93.68\% \\
%\hline
%4-layers& 78.50\%     & 94.27\%  &   93.68\% \\
%\hline
%4-layers(little)& 78.50\%     & 94.27\%  &   93.68\% \\
%
%\hline
%\end{tabular}
%\end{center}
%\vspace*{-0.5cm}
%\end{table}

\begin{table}[!htb]
\caption{Classification accuracies of adaptive routing algorithm in different numbers of capsule layers and different values of $\lambda$ on the dataset Fashion-MNIST\cite{fashionmnist}.}
\begin{center}
\label{result2}
\begin{tabular}{|l|c|c|c|c|c|}
\hline
   & \ $\lambda$=1 & \ $\lambda$=2 & \ $\lambda$=3  & \ $\lambda$=4\\
\hline
2-layers&  92.78\%     & 93.23\%  &   93.07\% &   92.96\%\\
\hline
3-layers& 93.54\%     & 93.63\%  &   93.39\% &   93.38\%\\
\hline
4-layers& 93.61\%     & \bf93.71\%  &   93.57\% &   93.41\%\\
\hline
\end{tabular}
\end{center}
\vspace*{-0.5cm}
\end{table}

\begin{table}[!htb]
\caption{Classification accuracies of adaptive routing algorithm in different numbers of capsule layers and different values of $\lambda$ on the dataset CIFAR10\cite{cifar}.}
\begin{center}
\label{result3}
\begin{tabular}{|l|c|c|c|c|c|}
\hline
   & \ $\lambda$=1 & \ $\lambda$=2 & \ $\lambda$=3  & \ $\lambda$=4\\
\hline
2-layers&  78.24\%     & 77.97\%  &   78.41\% &   78.34\%\\
\hline
3-layers& 78.41\%     & 78.01\%  &   78.66\% &   78.44\%\\
\hline
4-layers& 78.42\%     & 78.13\%  &   \bf78.68\% &   78.50\%\\
\hline
\end{tabular}
\end{center}
\vspace*{-0.5cm}
\end{table}

From the Table \ref{result2} and \ref{result3} , we have obtained different performances in different numbers of capsule layers and different values of $\lambda$ on the dataset CIFAR10\cite{cifar} and Fashion-MNIST\cite{fashionmnist}. When the other configuration parameters are identical, the performance of the model improved with the number of capsule layers increased. Moreover, the performance of the model is different by $\lambda$. When the value of $\lambda$ is 2 or 3, the performance is better on the the dataset Fashion-MNIST. And when the value of $\lambda$ equals to 1 or 3, we can also obtain the better performance on the dataset CIFAR10.

\begin{table}[!htb]
\caption{Classification accuracies of adaptive routing algorithm in small values of $\lambda$ on the dataset CIFAR10\cite{cifar}.}
\begin{center}
\label{result4}
\begin{tabular}{|l|c|c|c|c|c|}
\hline
   & \ $\lambda$=0.1 & \ $\lambda$=0.001 & \ $\lambda$=0.0001&$\lambda$=0.00001  \\
\hline
2-layers&  77.24\%     & 69.25\%  & 10.58\%  &    10.42\%\\
\hline
3-layers& 10.23\%     & 10.01\%  &  10.22\%  &   10.12\%\\
\hline
4-layers& 10.18\%     & 10.15\%  &  10.02\%  &   10.06\%\\
\hline
\end{tabular}
\end{center}
\vspace*{-0.5cm}
\end{table}

From the Table \ref{result4}, we have obtained different performances in small values of $\lambda$ on the dataset CIFAR10\cite{cifar}. It is obvious that there are two situations leading to the capsule networks not working, First,  the capsule network will collapse when the value of $\lambda$ is setted to 0.0001 or even less in two capsule layers which is same as the original paper\cite{DBLP:conf/nips/SabourFH17}. Second, when the value of $\lambda$ is setted to 0.1 or even less in multiple capsule layers (3-layers and 4-layers), the capsule network is not working too. Also, capsule networks using dynamic routing algorithm has the same situation when stacking multiple capsule layers. In the end, by comparing the results of multiple capsule layers in Table \ref{result3} and Table \ref{result4}, it proved that too small gradient coefficients in the capsule network result in the gradient vanishing and according to the value of the coupling coefficient $c_{ij}$ in Figure \ref{P1}

 In our proposed algorithm, $\lambda$ is equivalent to the number of iterations in the routing algorithm. In the capsule network, although the number increasing of iterations brings noise,  it can enhance the activation probability of high-layer capsules.  Further, we can get the best performance in the original capsule network when the number of iterations is three. In the end, although the meaning of the hyper-parameter $\lambda$ is the same as the number of iterations, the scale is different.

%So when $\lambda$ is 2 or 3, the performance of the model will be better.

%\begin{table}[!htb]
%\caption{Classification accuracies of adaptive routing without iteration algorithm in different values of $\lambda$.}
%\begin{center}
%\label{result3}
%\begin{tabular}{|l|c|c|c|c|c|}
%\hline
%Value of $\lambda$  & \footnotesize{CIFAR10} & \footnotesize{SVHN} & \footnotesize{F-MNIST}\\
%\hline
%$\lambda$=1&  76.05\%     & 93.65\%  &   93.02\% \\
%\hline
%$\lambda$=2& 78.50\%     & 94.27\%  &   93.68\% \\
%\hline
%$\lambda$=3& 78.50\%     & 94.27\%  &   93.68\% \\
%\hline
%$\lambda$=3& 78.50\%     & 94.27\%  &   93.68\% \\
%
%\hline
%\end{tabular}
%\end{center}
%\vspace*{-0.5cm}
%\end{table}

%\subsection{Improved Performance With Capsule Networks}
%Because of the existence of the routing process, the gradient propagation between the capsule layers is affected by the traditional neural network propagation. The deep learning method to improve the network performance is to expand the depth and breadth of the network. For the capsule network, if there is Large enough memory, the number of capsules in each layer and the dimensions of the capsules are better expanded, but it is difficult to improve the depth of the network. To solve the normal propagation of the gradient and control the amount of calculation.

%---------------------------------------------------------------------
\section{Conclusion}
In the original capsule network(used dynamic routing algorithm), the gradient vanishes when the model stacks multiple capsules layers. We analyzed the forward and backward propagation of the data flow in the capsule network and found that the coupling coefficient $c_{ij}$ leads to the gradient vanishing. Therefore, we proposed the adaptive routing algorithm to overcome the disadvantage of gradient vanishing when the network stacks multiple capsule layers, which do not involve the coupling coefficient $c_{ij}$ in the routing process. Considering the process of routing iteration
will bring a large amount of computation, first, we derived the iterative process of the adaptive routing algorithm. Second, simplified the iteration of the routing by replacing the number of iteration with a hyper-parameter $\lambda$. The hyper-parameter $\lambda$ not only inhibits the gradient vanishing but also the appropriate $\lambda$ can magnify the gradient so that it can propagate more effectively to the front of the layers in the model. As a result, our proposed adaptive routing algorithm
can achieve better performance than Sabour \cite{DBLP:conf/nips/SabourFH17} on
Fashion-MNIST\cite{fashionmnist}, SVHN\cite{svhn} and CIFAR-10\cite{cifar}, and
have the state-of-the-art performance on MNIST\cite{mnist} datasets. Further, we have obtained different performance in the different numbers of capsule layers and different values of hyper-parameters $\lambda$ and analyzed the experimental results.

 As future work, we will continue to research the capsule network to increase the number of network layers while reducing the amount of computation.

%--------------------------------------------------------------------

{\small
\bibliographystyle{ieee_fullname}
\bibliography{capbib}

\begin{thebibliography}{10}\itemsep=-1pt

\bibitem{DBLP:conf/cvpr/DengDSLL009}
Jia Deng, Wei Dong, Richard Socher, Li{-}Jia Li, Kai Li, and Fei{-}Fei Li.
\newblock Imagenet: {A} large-scale hierarchical image database.
\newblock In {\em 2009 {IEEE} Computer Society Conference on Computer Vision
  and Pattern Recognition {(CVPR} 2009), 20-25 June 2009, Miami, Florida,
  {USA}}, pages 248--255, 2009.

\bibitem{DBLP:conf/ijcai/DingWGLW19}
Xinpeng Ding, Nannan Wang, Xinbo Gao, Jie Li, and Xiaoyu Wang.
\newblock Group reconstruction and max-pooling residual capsule network.
\newblock In {\em Proceedings of the Twenty-Eighth International Joint
  Conference on Artificial Intelligence, {IJCAI} 2019, Macao, China, August
  10-16, 2019}, pages 2237--2243, 2019.

\bibitem{DBLP:conf/cvpr/HeZRS16}
Kaiming He, Xiangyu Zhang, Shaoqing Ren, and Jian Sun.
\newblock Deep residual learning for image recognition.
\newblock In {\em 2016 {IEEE} Conference on Computer Vision and Pattern
  Recognition, {CVPR} 2016, Las Vegas, NV, USA, June 27-30, 2016}, pages
  770--778, 2016.

\bibitem{DBLP:conf/icann/HintonKW11}
Geoffrey~E. Hinton, Alex Krizhevsky, and Sida~D. Wang.
\newblock Transforming auto-encoders.
\newblock In {\em Artificial Neural Networks and Machine Learning - {ICANN}
  2011 - 21st International Conference on Artificial Neural Networks, Espoo,
  Finland, June 14-17, 2011, Proceedings, Part {I}}, pages 44--51, 2011.

\bibitem{DBLP:conf/iclr/HintonSF18}
Geoffrey~E. Hinton, Sara Sabour, and Nicholas Frosst.
\newblock Matrix capsules with {EM} routing.
\newblock In {\em 6th International Conference on Learning Representations,
  {ICLR} 2018, Vancouver, BC, Canada, April 30 - May 3, 2018, Conference Track
  Proceedings}, 2018.

\bibitem{DBLP:conf/nips/JaderbergSZK15}
Max Jaderberg, Karen Simonyan, Andrew Zisserman, and Koray Kavukcuoglu.
\newblock Spatial transformer networks.
\newblock In {\em Advances in Neural Information Processing Systems 28: Annual
  Conference on Neural Information Processing Systems 2015, December 7-12,
  2015, Montreal, Quebec, Canada}, pages 2017--2025, 2015.

\bibitem{DBLP:conf/icml/JeongLK19}
Taewon Jeong, Youngmin Lee, and Heeyoung Kim.
\newblock Ladder capsule network.
\newblock In {\em Proceedings of the 36th International Conference on Machine
  Learning, {ICML} 2019, 9-15 June 2019, Long Beach, California, {USA}}, pages
  3071--3079, 2019.

\bibitem{DBLP:journals/corr/KingmaB14}
Diederik~P. Kingma and Jimmy Ba.
\newblock Adam: {A} method for stochastic optimization.
\newblock In {\em 3rd International Conference on Learning Representations,
  {ICLR} 2015, San Diego, CA, USA, May 7-9, 2015, Conference Track
  Proceedings}, 2015.

\bibitem{cifar}
Alex Krizhevsky, Geoffrey Hinton, et~al.
\newblock Learning multiple layers of features from tiny images.
\newblock Technical report, Citeseer, 2009.

\bibitem{mnist}
Yann LeCun, L{\'e}on Bottou, Yoshua Bengio, Patrick Haffner, et~al.
\newblock Gradient-based learning applied to document recognition.
\newblock {\em Proceedings of the IEEE}, 86(11):2278--2324, 1998.

\bibitem{DBLP:conf/nips/LenssenFL18}
Jan~Eric Lenssen, Matthias Fey, and Pascal Libuschewski.
\newblock Group equivariant capsule networks.
\newblock In {\em Advances in Neural Information Processing Systems 31: Annual
  Conference on Neural Information Processing Systems 2018, NeurIPS 2018, 3-8
  December 2018, Montr{\'{e}}al, Canada.}, pages 8858--8867, 2018.

\bibitem{DBLP:conf/eccv/LiGDOW18}
Hongyang Li, Xiaoyang Guo, Bo Dai, Wanli Ouyang, and Xiaogang Wang.
\newblock Neural network encapsulation.
\newblock In {\em Computer Vision - {ECCV} 2018 - 15th European Conference,
  Munich, Germany, September 8-14, 2018, Proceedings, Part {XI}}, pages
  266--282, 2018.

\bibitem{svhn}
Yuval Netzer, Tao Wang, Adam Coates, Alessandro Bissacco, Bo Wu, and Andrew~Y
  Ng.
\newblock Reading digits in natural images with unsupervised feature learning.
\newblock In {\em Neural Information Processing Systems Workshop(NeurIPSW)},
  2011.

\bibitem{DBLP:conf/cvpr/RajasegaranJJJS19}
Jathushan Rajasegaran, Vinoj Jayasundara, Sandaru Jayasekara, Hirunima
  Jayasekara, Suranga Seneviratne, and Ranga Rodrigo.
\newblock Deepcaps: Going deeper with capsule networks.
\newblock In {\em {IEEE} Conference on Computer Vision and Pattern Recognition,
  {CVPR} 2019, Long Beach, CA, USA, June 16-20, 2019}, pages 10725--10733,
  2019.

\bibitem{DBLP:conf/nips/SabourFH17}
Sara Sabour, Nicholas Frosst, and Geoffrey~E. Hinton.
\newblock Dynamic routing between capsules.
\newblock In {\em Advances in Neural Information Processing Systems 30: Annual
  Conference on Neural Information Processing Systems 2017, 4-9 December 2017,
  Long Beach, CA, {USA}}, pages 3856--3866, 2017.

\bibitem{DBLP:journals/corr/SimonyanZ14a}
Karen Simonyan and Andrew Zisserman.
\newblock Very deep convolutional networks for large-scale image recognition.
\newblock In {\em 3rd International Conference on Learning Representations,
  {ICLR} 2015, San Diego, CA, USA, May 7-9, 2015, Conference Track
  Proceedings}, 2015.

\bibitem{DBLP:conf/cvpr/SzegedyLJSRAEVR15}
Christian Szegedy, Wei Liu, Yangqing Jia, Pierre Sermanet, Scott~E. Reed,
  Dragomir Anguelov, Dumitru Erhan, Vincent Vanhoucke, and Andrew Rabinovich.
\newblock Going deeper with convolutions.
\newblock In {\em {IEEE} Conference on Computer Vision and Pattern Recognition,
  {CVPR} 2015, Boston, MA, USA, June 7-12, 2015}, pages 1--9, 2015.

\bibitem{fashionmnist}
Han Xiao, Kashif Rasul, and Roland Vollgraf.
\newblock Fashion-mnist: a novel image dataset for benchmarking machine
  learning algorithms.
\newblock {\em CoRR}, abs/1708.07747, 2017.

\bibitem{DBLP:conf/nips/ZhangEQ18}
Liheng Zhang, Marzieh Edraki, and Guo{-}Jun Qi.
\newblock Cappronet: Deep feature learning via orthogonal projections onto
  capsule subspaces.
\newblock In {\em Advances in Neural Information Processing Systems 31: Annual
  Conference on Neural Information Processing Systems 2018, NeurIPS 2018, 3-8
  December 2018, Montr{\'{e}}al, Canada.}, pages 5819--5828, 2018.

\end{thebibliography}
}
\end{document}